\documentclass[conference]{IEEEtran}
\IEEEoverridecommandlockouts
\usepackage{cite}
\usepackage{amsmath,amssymb,amsfonts}
\usepackage{algorithmic}
\usepackage{graphicx}
\usepackage{textcomp}
\usepackage{xcolor}
\usepackage{multirow}
\usepackage{booktabs}

\def\BibTeX{{\rm B\kern-.05em{\sc i\kern-.025em b}\kern-.08em
    T\kern-.1667em\lower.7ex\hbox{E}\kern-.125emX}}
\begin{document}

\title{Perception Reinforcement Using Auxiliary Learning Feature Fusion: A Modified Yolov8 for Head Detection \\}
\author{
\IEEEauthorblockN{1\textsuperscript{st} Jiezhou Chen}
\IEEEauthorblockA{\textit{College of Mechatronics and Control} \\
\textit{Engineering} \\
\textit{Shenzhen University}\\
Shenzhen, China \\
chenjiezhou2022@email.szu.edu.cn}
\and
\IEEEauthorblockN{2\textsuperscript{nd} Guankun Wang}
\IEEEauthorblockA{\textit{Department of Electronic Engineering} \\
\textit{The Chinese University of Hong Kong}\\
Hong Kong, China \\
gkwang@link.cuhk.edu.hk}
\and
\IEEEauthorblockN{3\textsuperscript{rd} Weixiang Liu}
\IEEEauthorblockA{\textit{College of Mechatronics and Control} \\
\textit{Engineering} \\
\textit{Shenzhen University}\\
Shenzhen, China \\
wxliu@szu.edu.cn} \\
\and
\IEEEauthorblockN{4\textsuperscript{th} Xiaopin Zhong}
\IEEEauthorblockA{\textit{College of Mechatronics and Control} \\
\textit{Engineering} \\
\textit{Shenzhen University}\\
Shenzhen, China \\
xzhong@szu.edu.cn}
\and
\IEEEauthorblockN{5\textsuperscript{th} Yibin Tian}
\IEEEauthorblockA{\textit{College of Mechatronics and Control} \\
\textit{Engineering} \\
\textit{Shenzhen University}\\
Shenzhen, China \\
ybtian@szu.edu.cn}
\and
\IEEEauthorblockN{6\textsuperscript{th} ZongZe Wu}
\IEEEauthorblockA{\textit{College of Mechatronics and Control} \\
\textit{Engineering} \\
\textit{Shenzhen University}\\
Shenzhen, China \\
zzwu@szu.edu.cn}}

\maketitle

\begin{abstract}
Head detection provides distribution information of pedestrian, which is crucial for scene statistical analysis, traffic management, and risk assessment and early warning. However, scene complexity and large-scale variation in the real world make accurate detection more difficult. Therefore, we present a modified Yolov8 which improves head detection performance through reinforcing target perception. An Auxiliary Learning Feature Fusion (ALFF) module comprised of LSTM and convolutional blocks is used as the auxiliary task to help the model perceive targets. In addition, we introduce Noise Calibration into Distribution Focal Loss to facilitate model fitting and improve the accuracy of detection. Considering the requirements of high accuracy and speed for the head detection task, our method is adapted with two kinds of backbone, namely Yolov8n and Yolov8m. The results demonstrate the superior performance of our approach in improving detection accuracy and robustness.
\end{abstract}


\section{Introduction}
Object detection has become a hot topic in computer vision tasks, which involves detecting instances of certain types of visual objects (e.g., humans, animals, or cars) in digital images~\cite{zou2023object}. Human head detection is one of the key problems in public safety scene understanding. Its goal is to estimate the distribution information of pedestrians in the scene, which plays an important role in risk perception and early warning, traffic control and scene statistical analysis. If pedestrian distribution statistics can be shown in real-time, the staff can be more targeted for statistical analysis, order maintenance and dangerous prevention~\cite{tripathi2019convolutional}.

 Due to scene complexity, large-scale variance, and high crowd density in the real world, head detection remains a challenging problem even though detection models have made considerable strides in recent years. Among many affecting factors, large-scale transformations and changes in the number of targets due to drastic variations of the scenes have a deleterious effect on the performance of the detection models. To tackle the problem of scene variations, many researchers try to take advantage of multi-scale features generated in deep convolutional neural networks at different levels~\cite{peng2018detecting, girshick2015fast}. Feature pyramid network (FPN) is the typical architecture for multi-scale feature fusion. However, its performance suffers noticeably when detecting dense and small objects. To address these issues, we attempt to enhance the detector's perception of the human head by introducing feature fusion through auxiliary learning.

\begin{figure*}
    \centering
    \includegraphics[width=0.9\textwidth]{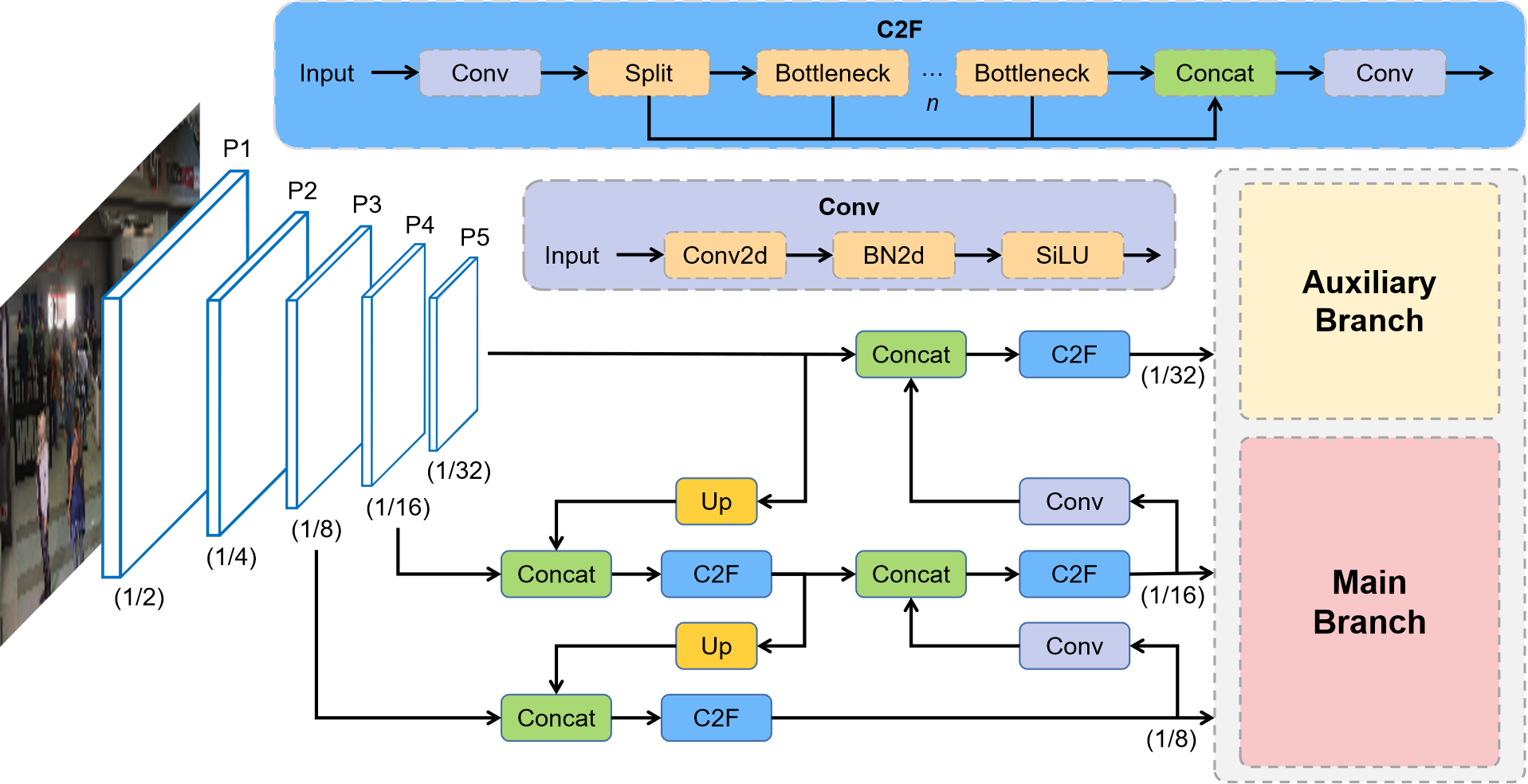}
    \caption{Overview of our proposed network based on Yolov8. It contains two branches which share the feature map from the Neck. We construst the ALFF module in Auxiliary branch. NC-DFL is employed as the optimization function in the main branch.}
    \label{fig: main}
\end{figure*}

Since the excellent detection performance of Yolov8, we make it as our baseline model. Our auxiliary learning method concentrates after the Neck part and introduces a heatmap prediction task that treats human head as positive class and the other areas as negative class. The heatmap prediction task is responsible for estimating the locations of the object centers. The response value in the heatmap is expected to be 1 when it locates in the center of the ground-truth object and the value gradually decreases exponentially as the distance between the heatmap location and the object center increases. The largest feature map is shared by the Yolov8 main task and the auxiliary learning task. Therefore, an additional reward from the auxiliary learning task can assist the main detection task to better perceive the position and size of the center of the human head. Consequently, models that incorporate an auxiliary learning task are able to superior detection performance compared to models that lack this task. In order to make full use of the features from main branch and conduct feature fusion, long short-term memory (LSTM)~\cite{hochreiter1997long} is employed as a bridge between convolutional blocks. By leveraging the dependencies among the features, LSTM effectively filters out redundant information. Furthermore, the framework employs our proposed Noisy Calibration Distribution Focal Loss (NC-DFL) to promote model fitting and improve the accuracy and robustness of detection.

\section{Related work}

\subsection{Head Detection}
Head detection methods can be categorized into traditional methods and deep learning methods. The traditional methods work by using handcrafted features ~\cite{enzweiler2008monocular, lin2010shape}, which suffers from limitations due to poor ability to generalise about the overall or local appearance of a person. With the advancements in deep learning, an increasing number of researchers are focusing their efforts on head detection~\cite{han2023cfnet, liu2021head, li2019headnet}. Currently, there are two types of deep learning-based object detection: one-stage detection and two-stage detection. The most representative two-stage detection method is Region-CNN (RCNN)~\cite{zhang2018occlusion}. Many different RCNN variations, including Faster RCNN~\cite{ren2015faster} and Mask RCNN~\cite{he2017mask}, have been used to count crowds in recent years. The efficiency of one-stage detectors is significantly higher but the accuracy is lower because they directly predict the bounding box of targets without region proposal.\cite{zhang2021vit} proposes Yolo network, which utilizes DarkNet as the feature extraction method. This allows Yolo to perform end-to-end optimization training, enabling it to simultaneously predict object bounding box locations and category probabilities. Since then many excellent one-stage detection models have been proposed, such as RetinaNet~\cite{lin2017focal}, CenterNet~\cite{zhou2019objects} and FCOS~\cite{tian2019fcos}.

\subsection{Auxiliary Learning}
Auxiliary learning is a technique that enhances the generalization ability of a primary task to handle unseen data, through training the primary task alongside auxiliary tasks~\cite{liu2019self}. By sharing features across these tasks, the model acquires relevant features that would not have been learned if it had only been trained on the main task. This broader support of features improves generalization to the main task even in the case of different interpretations of the input data. While similar to multi-task learning~\cite{caruana1997multitask}, auxiliary learning differs in that the primary task is performed with priority, while the auxiliary task is designed to assist the primary task. 
Auxiliary learning has demonstrated promising results in various domains, such as reinforcement learning~\cite{jaderberg2016reinforcement}, visual localization~\cite{valada2018deep}, and Instrument Segmentation~\cite{islam2019real}. To our knowledge, we are the first to apply auxiliary learning to head detection. 


\section{Methodology}
In this paper, Our approach reinforces the target perception capability of Yolov8 by introducing Auxiliary Learning Feature Fusion (ALFF). The overview network can be referred to in Figure~\ref{fig: main}. Besides, we also introduce NC-DFL which facilitates model parameter fitting.

\subsection{Network Architecture}
In Figure~\ref{fig: main}, the input of Yolov8 object detection network is 640x640 images and the backbone network and Neck are generally the same as the previous version. Based on the scaling factor, different size models of N/S/M/L/X scales are provided to meet different scenario requirements. To enrich the gradient flow, Yolov8 replaces the previous C3 module with the C2f module and adjusts the number of channels for different scale models. We introduce the auxiliary branch at the Head of Yolov8 so as to make full use of its excellent feature extraction capability.

\subsection{Auxiliary Learning Feature Fusion}
\subsubsection{Auxiliary Learning Task}    
\label{ALT}
\begin{figure}
    \centering
    \includegraphics[width=3.4in]{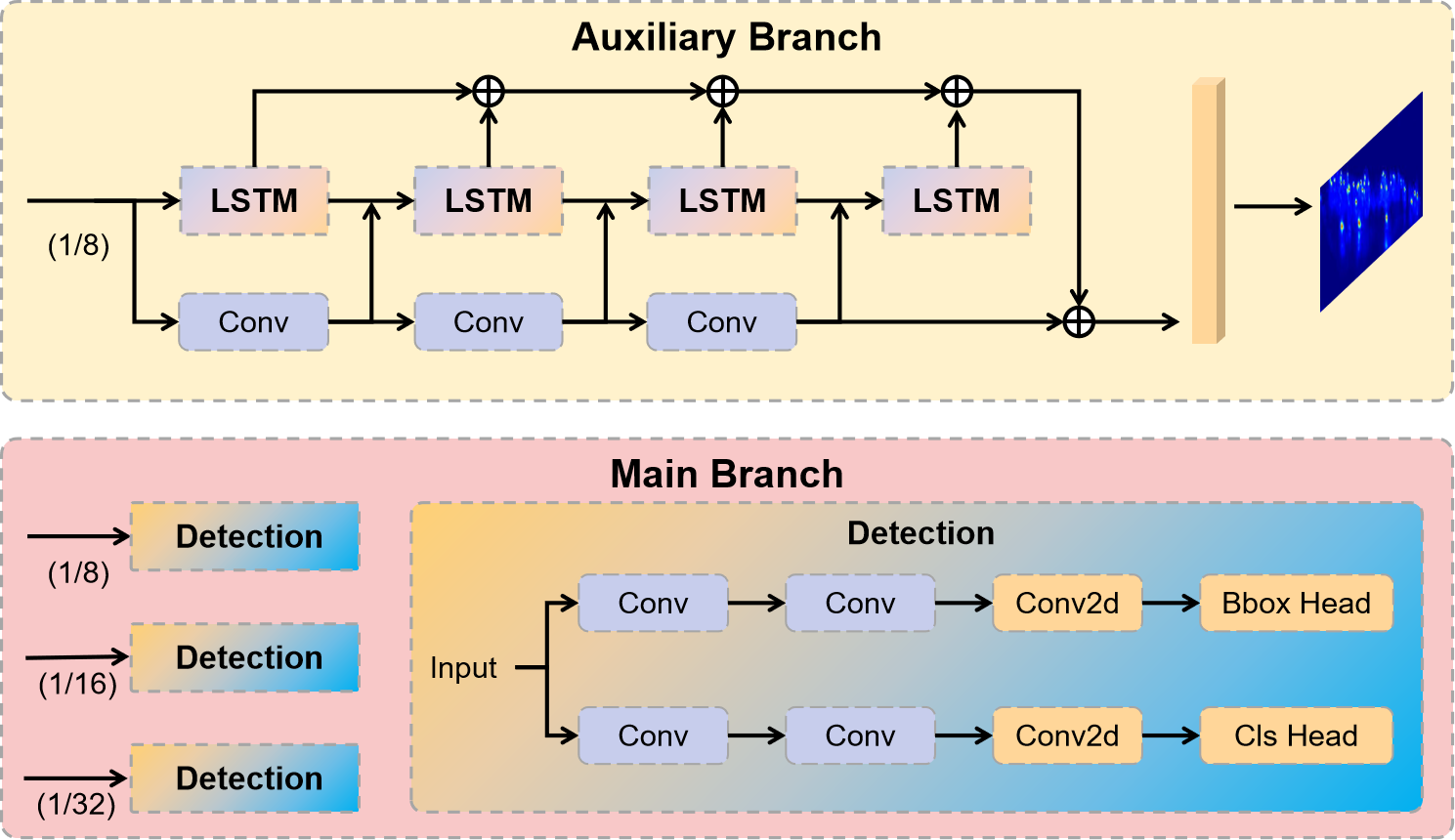}
    \caption{Branches of the modified Yolov8. The upper part is our proposed Auxiliary Learning Feature Fusion (ALFF) Module which utilizes LSTM and convolutional blocks for further feature refinement. Feature map of auxiliary branch (1/8) is shared with main branch.}
    \label{fig2}
\end{figure}

To detect multi-scale targets, Yolov8's Neck outputs features in three sizes including 1/8, 1/16 and 1/32. Since the feature map of size 1/8 preserves the most information among the output features, we shall use it as a shared representative for the main and auxiliary task. In the lower part of Figure~\ref{fig2}, the detection head of Yolov8 is replaced with the prevailing decoupled head construct, which separates the classification and detection heads. In addition, anchor-based is converted to anchor-free in order to enhance detection efficiency and accuracy. Therefore, encouraged by~\cite{zhou2019objects} which predicts the heatmap to estimate the location of targets, we set the auxiliary task as the prediction of human head heatmap. The auxiliary task is also responsible for estimating the locations of the object centers along with the main task, which can help the model better perceive the feature of targets.

For each bounding box $b^i = (c_x^i, c_y^i, w^i, h^i)$ in ground truth, the center of $i$th object is computed as $(c_x^i, c_y^i) = (\frac{x_1^i+x_2^i}{2}, \frac{y_1^i+y_2^i}{2})$ which will vary according to the size of the image scaling. Then, the heatmap response in $(x, y)$ can be defined as:
\begin{equation}
p_{xy}=\sum_{i=1}^{N} \exp ^{-{[(x-c_x^i)^2+(y-c_y^i)^2]}/{2\sigma^2}}
\end{equation}
where $N$ denotes the number of objects in the image and $\sigma$ represents the standard deviation. The scope of heatmap response is a circle of radius $min(w^i, h^i)$.

\subsubsection{Feature Fusion with LSTM}
As mentioned above, 1/8 scaling feature map contains abundant scale-relevant information. Subsequently, the ALFF is employed to enhance the features for more precise estimation. 
ALFF consists of an LSTM and three sequential convolutional blocks that serve as a feature refiner. The architecture of each convolutional block is shown in the \textit{Conv} of Figure~\ref{fig: main}. 

LSTM has been widely utilized in information processing tasks such as video processing and data systems. Its exceptional capability to capture long-term dependencies is essential. When it comes to image processing, a series of convolutional layers are typically employed to process features in a step-by-step manner. During this process, features exhibit similarity, correlation, and refinement, leading to strong dependencies between features at each level. Referring to the utility of LSTM in crowd counting~\cite{wu2023crowd}, we introduce LSTM into heatmap estimation as a information filter and dependency capturer. By leveraging the information from previous features, LSTM effectively filters the present feature.
In an LSTM module, each layer computes the input sequence with the following function:
\begin{equation}
    \begin{array}{l}
    i_{t}=\sigma\left(W_{i i} x_{t}+b_{i i}+W_{h i} h_{t-1}+b_{h i}\right) \\
    f_{t}=\sigma\left(W_{i f} x_{t}+b_{i f}+W_{h f} h_{t-1}+b_{h f}\right) \\
    g_{t}=\tanh \left(W_{i g} x_{t}+b_{i g}+W_{h g} h_{t-1}+b_{h g}\right) \\
    o_{t}=\sigma\left(W_{i o} x_{t}+b_{i o}+W_{h o} h_{t-1}+b_{h o}\right) \\
    c_{t}=f_{t} \odot c_{t-1}+i_{t} \odot g_{t} \\
    h_{t}=o_{t} \odot \tanh \left(c_{t}\right)
\end{array}
\end{equation}
where $i_t$, $f_t$, $g_t$, $o_t$ are the input, forget, cell, and output gates, respectively. $h_t$ is the hidden state at time \textit{t}, $c_t$ is the cell state at time \textit{t}, $x_t$ is the input at time \textit{t}, and $h_{t-1}$ is the hidden state of the layer at time \textit{t-1} or the initial hidden state at time 0. $\sigma$ is the sigmoid function, and  $\odot$  is the Hadamard product. $W_p$ and $b_p$ denote the weight matrix and the bias term of the gate \textit{p}.
 
In ALFF, the LSTM connects sequential features by making features as multi-node inputs and thus, which takes full advantage of the dependencies between hierarchical features for fusion. The LSTM is designed to use dependencies and filter hierarchical features. It employs input, memory, forget, and output gates to control the transmission state. The transmission state decides the filtering of the input feature, enabling the LSTM cell to produce the output that preserves valuable information while filtering out redundancy. Additionally, information from the current LSTM cell is passed on to the following cell, assisting it in filtering the subsequent input feature. By effectively concatenating the LSTM outputs, the information on hierarchical features can be well preserved. At the end of the auxiliary branch, a fully connected layer is used to modify the number of channels, followed by an upsampling layer that adjusts the feature map to the original size of the input.

\subsection{Noisy Calibration Distribution Focal Loss}
There are significant differences in scene and target distributions between the different datasets of head detection. When the number of targets increases or the scene changes drastically, it is difficult for the model to predict the location and size of the targets accurately. We aim to modulate bounding box loss function to produce a stable and robust training procedure for the predictor to handle different datasets. The bounding box loss function utilized by Yolov8 is Distribution Focal Loss~\cite{li2022generalized} which can be described as:
\begin{equation}
\begin{aligned}
\operatorname{DFL}\left(P_{i}, P_{i+1}\right)=-\left(\left(y_{i+1}-y\right) \log \left(P_{i}\right) \right. \\  \left. +\left(y-y_{i}\right) \log \left(P_{i+1}\right)\right)
\end{aligned}
\end{equation}

where $y_i$ and $y_{i+1}$ is explicitly enlarged probability (the nearest two to $y$). $P_{i}$ is short for $P(y_i)$ which is the probability distribution of $y_i$.

When the output probabilities closely align with the desired outcome, it can be defined as individual saturation. However, early saturation will limit the optimizer to move. As the number of saturated individuals increases, it is more likely to converge to a local minima, leading to overfitting eventually~\cite{chen2017noisy}. To mitigate this, it is beneficial to appropriately delay sample saturation to promote global convergence in regular experiments. For datasets with simple scenarios and fewer targets, the above procedure is effective. However, achieving convergence is particularly challenging when the scene is complex and the number of targets is large. During the training stage, the detection model remains underfitted for most samples. To address this, we adopt an opposite perspective and promote sample saturation. By increasing the logit value of underfitted samples, we enhance their contribution during back-propagation. This adjustment provides the optimizer with more opportunities to progress toward global convergence. To calibrate the logit values, we utilize Gaussian noise as a carrier, effectively aiding model fitting. Therefore, the logits fed into noise calibration can be rewritten as:

\begin{equation}
y^G=y[1+{\alpha}\cdot(\mu+\sigma\left|\xi\right|)]
\end{equation}

where $\mu$ and $\sigma$ are the mean and the standard variance of the Gaussian noise. $\xi$ $\sim$ $\mathcal{N}$(0,1). hyper-parameter $\alpha$ is used to adjust the scale of noise. Here, we set the mean value to 0. Therefore, the model can adjust the logits and get better global convergence. Gaussian noise can further improve the robustness of training effectively. 

In contrast, our NC-DFL can be calibrated for a simple training set as follows

\begin{equation}
y^G=y[1-{\alpha}\cdot(\mu+\sigma\left|\xi\right|)]
\end{equation}

We hereby obtain our noisy calibration distribution focal loss (NC-DFL):

\begin{equation}
\begin{aligned}
\operatorname{DFL}\left(P_{i}, P_{i+1}\right)^G=-\left(\left(y^G_{i+1}-y^G\right) \log \left(P^G_{i}\right) \right. \\  \left. +\left(y^G-y^G_{i}\right) \log \left(P^G_{i+1}\right)\right),\\
y^G=y[1\pm {\alpha}\cdot(\mu+\sigma\left|\xi\right|)]
\end{aligned}
\end{equation}

\section{Experiments}

\subsection{Datasets and Settings}
\begin{figure}[!ht]
    \centering
    \includegraphics[width=0.48\textwidth]{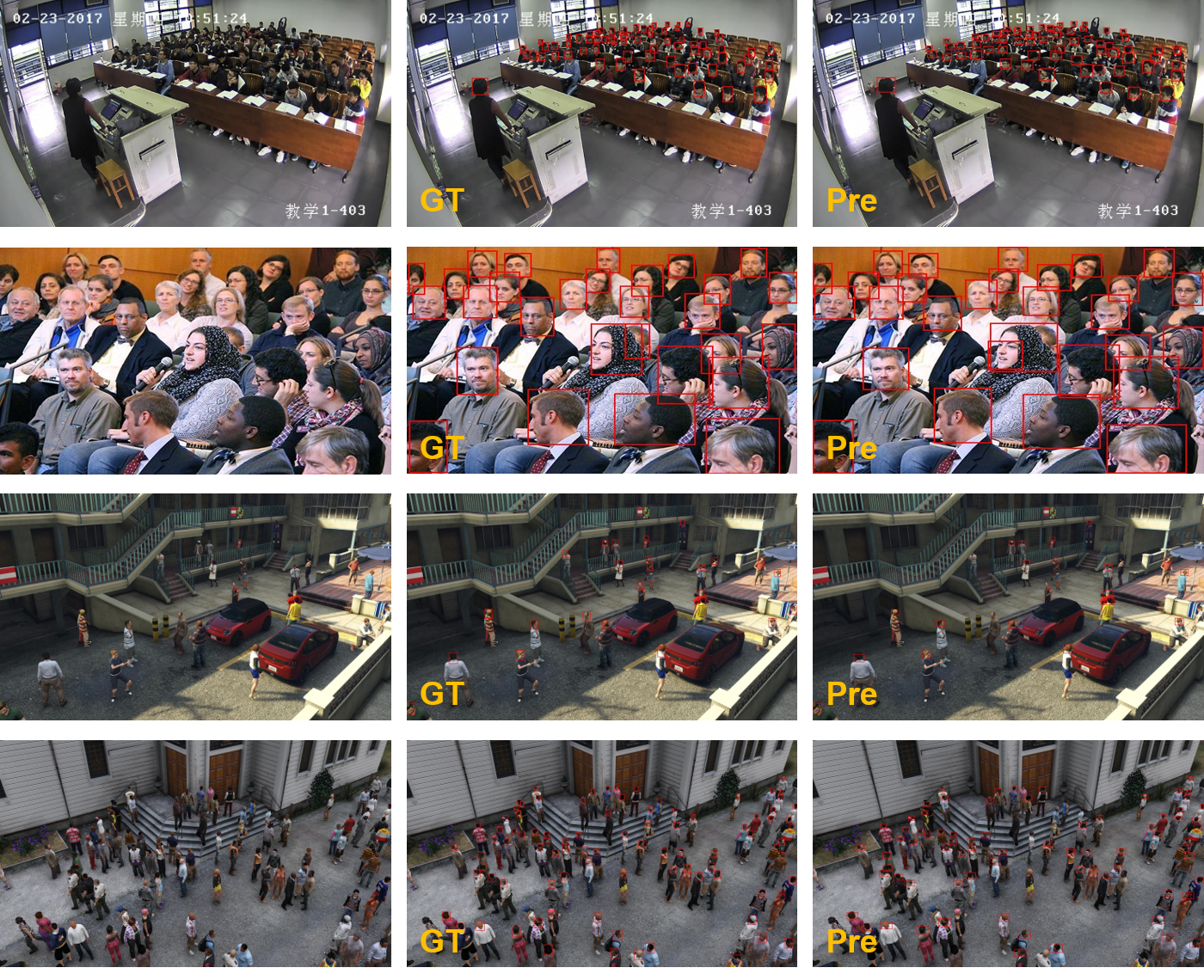}
    \caption{From top to bottom are exmples from SCUT-HEAD PartA, SCUT-HEAD PartB, GTA\_L and GTA\_H.}
    \label{fig:example}
\end{figure}

We evaluate our proposed method on two public datasets. The detection results of both datasets are visualized in Figure~\ref{fig:example}.

\textbf{SCUT-HEAD:} SCUT-HEAD~\cite{peng2018detecting} is a publicly large-scale head detection dataset, including 4405 frames labeled with 111251 bounding boxes. There are two parts in this dataset. The first part consists of 2,000 images extracted from classroom surveillance video. The other part consists of 2,405 images captured from the Internet.

\textbf{GTA\_Head:} GTA\_Head~\cite{zhong2022mask} is a large-scale virtual world dataset for head detection, including 5096 images labeled with 1732043 bounding boxes. It contains numerous complex scenes: stadiums, indoor shopping malls, subways and squares, etc. To assess our framework under different pedestrian densities, we select scenes according to the density and divide them into GTA\_L and GTA\_H based on the average number of person. Those scenes with density of less than 100 per frame are covered in GTA\_L which  consists of a total of 8 scenarios, with 5 scenarios in the training set and 3 scenarios in the testing set. The GTA\_H comprises 8 scenes in the training set and 4 scenes in the testing set and the density greater than 100 less than 300. The scene distribution of GTA\_L and GTA\_H can be found in Table~\ref{tab:gtadata}.

\begin{table}[]
\caption{Snece distribution of GTA\_L and GTA\_H datasets in GTA\_Head.}
\label{tab:gtadata}
\centering
\renewcommand{\arraystretch}{1.1}
\begin{tabular}{ccccc}
\toprule[1pt]
\multirow{2}{*}{Density}                                                & \multicolumn{2}{c}{Training Set}                        & \multicolumn{2}{c}{Testing Set}                     \\ \cline{2-5} 
                                                                        & \multicolumn{1}{l}{Snece} & \multicolumn{1}{l}{Number/Image}     & \multicolumn{1}{l}{Snece} & \multicolumn{1}{l}{Number/Image} \\ \hline
\multirow{5}{*}{\begin{tabular}[c]{@{}c@{}}Low\\ Density\end{tabular}}  & 7                         & \multicolumn{1}{c|}{77.30}  &                           &                         \\
                                                                        & 9                         & \multicolumn{1}{c|}{93.97}  & 33                        & 34.69                   \\
                                                                        & 17                        & \multicolumn{1}{c|}{41.95}  & 34                        & 41.99                   \\
                                                                        & 19                        & \multicolumn{1}{c|}{76.73}  & 35                        & 28.25                   \\
                                                                        & 20                        & \multicolumn{1}{c|}{39.25}  &                           &                         \\ \hline
\multirow{8}{*}{\begin{tabular}[c]{@{}c@{}}High\\ Density\end{tabular}} & 3                         & \multicolumn{1}{c|}{256.68} &                           &                         \\
                                                                        & 4                         & \multicolumn{1}{c|}{224.31} &                           &                         \\
                                                                        & 5                         & \multicolumn{1}{c|}{284.10} & 22                        & 228.01                  \\
                                                                        & 8                         & \multicolumn{1}{c|}{105.99} & 24                        & 127.11                  \\
                                                                        & 11                        & \multicolumn{1}{c|}{235.22} & 27                        & 159.01                  \\
                                                                        & 12                        & \multicolumn{1}{c|}{185.99} & 29                        & 216.69                  \\
                                                                        & 13                        & \multicolumn{1}{c|}{195.80} &                           &                         \\
                                                                        & 21                        & \multicolumn{1}{c|}{286.26} &                           &                         \\
\bottomrule[1pt]
\end{tabular}
\end{table}

\subsection{Implementation details}

There are different variants with distinct network architectures in Yolov8. Considering the requirements of high accuracy and speed for the head detection task, we adopt Yolov8n and Yolov8m pretrained on COCO as our representative baseline models. In the following experiments, all methods are implemented using the Python Pytorch framework on an NVIDIA RTX 3090Ti GPU. The batch size is set to 16 for 50 epochs. We adopt the initial learning rate $1 \times 10^{-2}$ with the weight decay of $5 \times 10^{-4}$. Based on our pre-training results, we find that $\alpha$ in Equation 6 works best when it is 1.0, which will be maintained in subsequent experiments.

\subsection{Results and Discussion}

\begin{table*}[]
\caption{The detection results of the proposed method with other classical methods on SCUT-HEAD datasets.}
\label{tab:scut}
\centering
\renewcommand{\arraystretch}{1.2}
\setlength{\tabcolsep}{5.7mm}{
\begin{tabular}{cccccccc}
\toprule[1pt]
\multirow{2}{*}{Approach} &   \multirow{2}{*}{Backbone}                            & \multicolumn{3}{c}{SCUT-HEAD\_PartA}         & \multicolumn{3}{c}{SCUT-HEAD\_PartB} \\ \cline{3-8}
                          &                       & AP50   & AP75   & AP50-95                     & AP50       & AP75       & AP50-95     \\ \hline
RetinaNet~\cite{lin2017focal}                 & \multicolumn{1}{c|}{R50-FPN}  & 0.904  & 0.203  & \multicolumn{1}{c|}{0.386}  & 0.868      & 0.512      & 0.489       \\
CenterNet~\cite{zhou2019objects}                 & \multicolumn{1}{c|}{R50-FPN}  & 0.949  & \textbf{0.414}  & \multicolumn{1}{c|}{0.476}  & 0.937      & 0.573      & 0.54        \\
FCOS~\cite{tian2019fcos}                      & \multicolumn{1}{c|}{R101-FPN} & 0.927  & 0.405  & \multicolumn{1}{c|}{0.471}  & 0.898      & 0.487      & 0.456       \\
Foveabox~\cite{kong2020foveabox}                  & \multicolumn{1}{c|}{R101}     & 0.9    & 0.194  & \multicolumn{1}{c|}{0.377}  & 0.894      & 0.485      & 0.482       \\
Yolov8                    & \multicolumn{1}{c|}{YOLOv5n}  & 0.9503 & 0.2837 & \multicolumn{1}{c|}{0.4371} & 0.9492     & 0.6027     & 0.5561      \\
Yolov8(Ours)       & \multicolumn{1}{c|}{YOLOv5n}  & 0.95   & 0.2958 & \multicolumn{1}{c|}{0.441}  & 0.9496     & 0.6053     & 0.558       \\
Yolov8                    & \multicolumn{1}{c|}{YOLOv5m}  & 0.9703 & 0.332  & \multicolumn{1}{c|}{0.4666} & 0.9624     & 0.6285     & 0.5750      \\
Yolov8(Ours)       & \multicolumn{1}{c|}{YOLOv5m}  & \textbf{0.9715} & 0.3682 & \multicolumn{1}{c|}{\textbf{0.4808}} & \textbf{0.9633}     & \textbf{0.6299}     & \textbf{0.5752}      \\ \bottomrule[1pt]
\end{tabular}}
\end{table*}

\begin{table*}[]
\caption{The detection results of the proposed method with other classical methods on GTA\_L and GTA\_H datasets.}
\label{tab:gta}
\centering
\renewcommand{\arraystretch}{1.2}
\setlength{\tabcolsep}{5.7mm}{
\begin{tabular}{cccccccc}
\toprule[1pt]
\multirow{2}{*}{Approach} & \multirow{2}{*}{Backbone}          & \multicolumn{3}{c}{GTA\_L}               & \multicolumn{3}{c}{GTA\_H} \\ \cline{3-8}
                          &                       & AP50   & AP75    & AP50-95                     & AP50      & AP75     & AP50-95   \\ \hline
RetinaNet~\cite{lin2017focal}                  & \multicolumn{1}{c|}{R50-FPN}  & 0.513  & 0.029   & \multicolumn{1}{c|}{0.127}  & 0.293     & 0.066    & 0.78      \\
CenterNet~\cite{zhou2019objects}                 & \multicolumn{1}{c|}{R50-FPN}  & 0.555  & 0.019   & \multicolumn{1}{c|}{0.144}  & 0.301     & 0.075    & 0.119     \\
FCOS~\cite{tian2019fcos}                       & \multicolumn{1}{c|}{R101-FPN} & 0.462  & 0.011   & \multicolumn{1}{c|}{0.107}  & 0.477     & 0.126    & 0.135     \\
Foveabox~\cite{kong2020foveabox}                     & \multicolumn{1}{c|}{R101}     & 0.521  & 0.023   & \multicolumn{1}{c|}{0.126}  & 0.439     & 0.101    & 0.125     \\
Yolov8                    & \multicolumn{1}{c|}{YOLOv5n}  & 0.5654 & 0.03641 & \multicolumn{1}{c|}{0.1616} & 0.4843    & 0.1527   & 0.2121    \\
Yolov8(Ours)       & \multicolumn{1}{c|}{YOLOv5n}  & 0.5725 & 0.0381  & \multicolumn{1}{c|}{0.1686} & 0.4849    & 0.1512   & 0.2104    \\
Yolov8                    & \multicolumn{1}{c|}{YOLOv5m}  & 0.6661 & 0.09988 & \multicolumn{1}{c|}{0.2133} & 0.5824    & 0.2143   & 0.2696    \\
Yolov8(Ours)       & \multicolumn{1}{c|}{YOLOv5m}  & \textbf{0.6907} & \textbf{0.1121}  & \multicolumn{1}{c|}{\textbf{0.2394}} & \textbf{0.591}     & \textbf{0.2292}   & \textbf{0.2808}    \\ \bottomrule[1pt]
\end{tabular}}
\end{table*}

\subsubsection{Evaluation on SCUT-HEAD}
Table~\ref{tab:scut} shows the quantitative experimental result of SCUT-HEAD dataset. Overall our modified Yolov8 achieved excellent performance and significantly outperforms other classical detectors in all evaluation metrics. As mentioned in Section 3, the introduction of the anchor-free prediction of human head heatmap auxiliary task aids the model in better perceiving the features of the targets. Consequently, both Yolov8 models are enhanced through our approach. Specifically, the modified Yolov8m performs best among all baselines with AP50 scores of 0.9715 and 0.9633 on two parts. Besides, for stricter metrics, our approach gives a 2.2\%-10.9\% improvement on AP75 and a 0.03\%-2.95\% improvement on AP50-95 for Yolov8 on both backbones.

\subsubsection{Evaluation on GTA\_Head}
The GTA\_L and GTA\_H datasets are more challenging because of the complex scenes and scale variations. Table~\ref{tab:gta} shows the quantitative experimental result. Yolov8m is also the best-performing detector, and our improvement method helps Yolov8m further improve the full range of evaluation metrics. More significant AP75 enhancement, with 12.23\% in GTA\_L and 6.95\% in GTA\_H. As evident from this, our improved Yolov8 utilizes the advanced model architecture of Yolov8, which not only performs well in simple scenes but also significantly enhances the accuracy of head detection in complex and high-density datasets. Moreover, our framework improves the perceptual capability of YOLOv8 while enhancing the robustness of the model across different scenes and target distributions. Due to the diverse nature of the GTA\_Head dataset, the detection performance of the original YOLOv8 decreases. Our NC-DFL method satisfactorily addresses this issue, achieving AP50 of 0.6907 in low-density scenarios and 0.591 in high-density scenarios within the GTA\_Head dataset. The superior performance of the modified Yolov8 in this challenging scenario further reflects the excellent universality and power of our method.

\begin{table}[]
\caption{Ablation study of our framework on the GTA\_H dataset. ALFF and NC-DFL denote auxiliary branch and Noisy Calibration Distribution Focal Loss. We remove the above components one by one and observe the detection performance.}
\label{tab:abl}
\renewcommand{\arraystretch}{1.2}
\centering
\setlength{\tabcolsep}{4mm}
\begin{tabular}{cc|ccc}
\toprule[1pt]
\multirow{2}{*}{ALFF}  & \multirow{2}{*}{NC-DFL}      & \multicolumn{3}{c}{GTA\_H} \\ \cline{3-5}
 &                    & AP50         & AP75         & AP50-95      \\ \hline
×    & × & 0.5824       & 0.2143       & 0.2696       \\
\checkmark    & × & 0.5866       & \textbf{0.2384}       & 0.2802       \\
×    & \checkmark & 0.5879       & 0.219        & 0.2742       \\
\checkmark    & \checkmark & \textbf{0.591}        & 0.2292       & \textbf{0.2808}     \\            
\toprule[1pt]
\end{tabular}
\end{table}

\subsubsection{Ablation Studies}
We make ablation study to assess the effects of removing ALFF and NC-DFL, respectively. For ablation dataset, we select the GTA\_H which presents the most challenging task. Experimental results in Table~\ref{tab:abl} conducted on the GTA\_H dataset reveal that in the absence of NC-DFL, our auxiliary branch with the LSTM network can improve the performance of head detection to some extent. Additionally, the standalone introduction of NC-DFL significantly enhances the performance of complex tasks. However, compared with our final results, we can observe the performance degradation of individual components. These observations indicate that ALFF and NC-DFL positively contribute to achieving the best results. 


\section{Conclusion}
In this paper, we propose a modified Yolov8 that enhances head detection performance by reinforcing target perception. To aid the model in perceiving targets, we introduce Auxiliary Learning Feature Fusion, which combines LSTM and convolutional blocks. This module provides additional information to enhance target understanding. Moreover, we leverage Noise Calibration in the Distribution Focal Loss to facilitate model fitting and improve the detection accuracy. Experimental results show the superior performance of our approach in Yolov8n and Yolov8m, leading to improved detection accuracy and robustness. In summary, our modified Yolov8 approach enhances head detection performance by employing ALFF and NC-DFL. In future work, we will simplify the model while maintaining accuracy and try to use our modified Yolov8 for the detection of a wider variety of targets.

\section{Acknowledgment}

This work was supported in part by Grants of National Key R\&D Program of China 2020AAA0108300, the Shenzhen University 2035 Program for Excellent Research 00000224 and  the National Nature Science Foundation of China under Grant 62171288.

\bibliographystyle{unsrt}
\bibliography{reference}

\end{document}